\PassOptionsToPackage{table,dvipsnames}{xcolor}

\documentclass[10pt, a4paper]{article}

\usepackage{lrec-coling2024} 

\usepackage[utf8]{inputenc}
\usepackage[russian,english]{babel}
\usepackage{hyperref}
\usepackage{todonotes}
\usepackage{subfig}
\usepackage{graphicx}
\usepackage{booktabs, multirow}
\usepackage{soul}
\usepackage{rotating}

\sethlcolor{lime}

\title{Little Red Riding Hood Goes Around the Globe: \\Crosslingual Story Planning and Generation with Large Language Models}

\name{Evgeniia Razumovskaia$^{1*}$\thanks{* The work was done while interning at Google.}, Joshua Maynez$^{2}$, Annie Louis$^{2}$, \\ {\bf \large Mirella Lapata$^{2}$, Shashi Narayan$^{2}$}} 

\address{$^{1}$ Language Technology Lab, University of Cambridge \\
         $^{2}$ Google DeepMind \\
         \texttt{er563@cam.ac.uk}, \texttt{joshuahm@google.com}, \texttt{annielouis@google.com}, \\
         \texttt{lapata@google.com}, \texttt{shashinarayan@google.com}\\}
         
\newcommand{\dataset}{\textsc{ASPEN}}

\abstract{
Previous work has demonstrated the effectiveness of planning for story
generation exclusively in a monolingual setting focusing primarily on
English.  We consider whether planning brings advantages to automatic
story generation across languages. We propose a new task of
crosslingual story generation with planning and present a new dataset for this task.  We conduct a comprehensive study of different plans and
generate stories in several languages, by leveraging the creative and
reasoning capabilities of large pretrained language models. Our
results demonstrate that plans which structure stories into three acts
lead to more coherent and interesting narratives, while allowing to explicitly control their content and structure. 
 \\ \newline  }

\begin{document}

\maketitleabstract

\section{Introduction}
\label{sec:Introduction}

Automated story generation has met with fascination since the early
days of artificial intelligence. Initial story generation systems
required substantial knowledge engineering to create symbolic domain
models that described legal characters and their actions, and have
almost ubiquitously relied on symbolic planning
\cite{Meehan:77,Lebowitz:ea:1987,Riedl:Young:2010,Ware:Young:2010,Cavazza:ea:2003,Liu:Singh:02}
and case-based reasoning
\cite{Perez:ea:2001,Peinado:Gervas:2005,Turner:92}. 

The advent of large pre-trained language models (PLMs;
\citealt{raffel2020exploring, lewis2020bart, brown2020language,
  chowdhery2022palm}) has provided a common framework for AI story
generation, eschewing the need for manual knowledge
engineering. Despite their ability to produce relatively fluent and
naturalistic text, language models struggle with maintaining story
coherence --- the logical progression of events --- and may also
become repetitive. Attempts to enhance the coherence of the 
stories and control the trajectory of events often decompose 
the generation task into \emph{planning} an outline or sketch, and then
\emph{elaborating} on it, e.g.,~by filling in descriptions, and
specific details of each story. Story plans have been previously
represented by keywords such as events or phrases
\cite{yao2019plan,xu-etal-2018-skeleton,rashkin-etal-2020-plotmachines},
a sequence of actions \cite{goldfarb2020content,fan2019strategies},
optionally combined with information on characters and setting
\cite{yang2022re3}, and control tokens
\cite{lin-riedl-2021-plug,peng-etal-2018-towards,ippolito-etal-2019-unsupervised,xu-etal-2020-megatron}.

Previous work has demonstrated the effectiveness of planning for story
generation exclusively in a monolingual setting focusing primarily on
English \cite{fang2021outline, alhussain2021automatic}.  We consider
whether planning brings advantages to automatic story generation
across languages. Specifically, we introduce a new task of
\textit{crosslingual} story generation: given a plan in English,
generate a coherent narrative in a target language which is consistent
with the contents of the plan. The task is challenging for several
reasons. Firstly, the plans provide only basic plot information about
the story. The model needs to flesh out these plot elements and
generate fluent text in the target language, while staying true to the
original plan. Moreover, there are differences in story telling
traditions amongst cultures and languages \cite{Zipes:2012}, which
felicitous stories meant for human readers would need to observe.

We generate stories in several languages leveraging the creative and
reasoning capabilities of pretrained language models
\cite{chowdhery2022palm,wei2022chain} which have recently achieved
strong performance on various tasks and in many languages
\citep{wei2022emergent, hao2022language, arora2022ask}. PLMs provide a
unified modeling framework for our task. Rather than building a
different model for language, the PLM is used to generate stories in
any target language, without further modification.  Following
\citet{brown2020language}, we consider a class of hand-designed
prompts referred to as ``prompting'' or ``in-context learning''. The
prompt starts with a free form instruction, followed by a small number
of instances exemplifying how the task is solved.  In our case,
prompts contain plans and their corresponding stories.


We further investigate which plans are best suited for crosslingual
story generation. Drawing inspiration from theories of discourse
structure \cite{Grosz:ea:95,Carlson:1983,roberts:2012:information} and
narrative fiction \cite{McKee:1997}, we create plan representations
which differ in form and content. As far as the content is concerned,
we formulate plans as a list of entities, a sequence of actions, and
events revolving around the three-act structure
\cite{field2005screenplay}, a popular story writing technique that
divides a narrative into three distinct parts: the setup, the
conflict, and the resolution. In terms of form, plans are keywords,
prosaic text, and question-answer pairs. To facilitate research in
this area, we further create a new dataset containing stories in 31
linguistically diverse languages collated from the Global African
Storybook
Project.\footnote{\href{https://global-asp.github.io/about/}{Global
    ASP} aims to translate African stories into all of the world's
  languages.}


Our results demonstrate that the three-act structure leads to better stories (across languages) compared to
less detailed alternatives. We hypothesize this is due to the inner
structure of the plan, i.e.,~the questions inform the model of important 
events (i.e.,~setup, plot, and resolution) and improve the
coherence of the story. In addition, our
work shows that large language models are capable of generating fluent
narratives in multiple languages, while mostly being pretrained on
English data. Our contributions can be summarized as follows: (a)~we propose a new
task of crosslingual story generation with planning and present a new
dataset\footnote{We will make the dataset available 
at \url{URL} to foster future work on crosslingual story planning and generation.} for this task; (b)~we conduct a comprehensive study of
different plans and prompt formulations for crosslingual story
generation; and (c)~present results, based on automatic and human assessments, which confirm that plans
following the three-act structure expressed as question-answer pairs
are more controllable and generate more coherent narratives.

\section{Related Work}
\label{sec:related-work}


\paragraph{Story Generation with Planning} Plans have been widely
used to improve coherence in automated story generation.  Several
approaches \cite{yao2019plan, fan2019strategies, goldfarb2020content,
  fang2021outline, yang2022re3} have relied on planning as an
intermediate step before generating the full story, with all narrative
details.  Intermediate plans have taken the form of a sequence of
entities and their actions \citep{yao2019plan}, outlines based on
semantic role labeling \cite{fan2019strategies}, plot structures
rooted in Aristotelian philosophy \cite{goldfarb2020content}, and more
elaborate descriptions including details about the setting of the
story, its characters, and main plot points \cite{yang2022re3}. These
efforts have consistently demonstrated that generating full stories
based on structured plans improves their coherence and overall
quality.  We build on this work in two ways: a)~we study plan-based
story generation in a crosslingual context, which, to our knowledge,
has not been done so far; and b)~we systematically study which type of
plan leads to better stories in this crosslingual setting. We simplify
the generation problem in that  plans are not learned or predicted
(e.g.,~\citealt{goldfarb2020content,yang2022re3}), instead they are
provided to the model to generate from. Future work could explore how
to generate plans in addition to stories given an initial
specification \cite{yang2022re3}.

Work on multilingual story generation has made little progress due to
scarcity of resources \citep{hou2019survey}. MTG \citep{chen2021mtg}
is a new multilingual benchmark covering several subtasks one of which
is story generation. Specifically, they translate ROCStories
\cite{mostafazadeh-etal-2016-corpus}, a widely used dataset for
testing performance on the Story Cloze task, into four languages
(\texttt{de, fr, es, zh}). We create the first crosslingual dataset
which allows to evaluate full stories (as opposed to story
completions) in a crosslingual setting.


\paragraph{Prompting}
Prompting aims to make better use of the knowledge encapsulated in
pretrained language models to solve various types of downstream tasks
by simply conditioning  on a few examples (few-shot) or instructions
describing the task at hand (zero-shot).  Existing work
\citep[][\emph{inter
  alia}]{brown2020language,schick-schutze-2021-just} has shown that
prompting can lead to strong performance in a wide range of tasks
including question answering \cite{chowdhery2022palm,Agrawal:ea:2022},
and open-ended natural language generation
\cite{tang-etal-2022-context}.  Prompting in multilingual settings has
achieved good performance using English prompts and target language
exemplars
\cite{winata-etal-2021-language,Lin:ea:2022,Shi:ea:2022}. Polyglot
prompting \citep{fu2022polyglot} aims to learn a unified semantic
space for different languages based on multilingual prompt
engineering.

Our work employs prompting in a crosslingual setting. We adopt a
unified modeling approach, we generate stories in different languages
with same model, by changing the language id and the story examples
presented to the model, while English plans remain fixed.


\section{The \dataset\ Benchmark for Crosslingual Story Generation}
\label{sec: Aspen}



\paragraph{Task Description}
Given a plan in English, our task is to generate a coherent narrative
in a target language which is consistent with the contents of the
plan. This renders our setting strictly crosslingual, i.e., we assume
models do not observe any target language text in the plan. We expect
plans to serve as structured information about story content, leaving
room for generating stories creatively in different languages.



\begin{table}[t]
\footnotesize
\resizebox{\columnwidth}{!}{\begin{tabular}{@{}l@{~}|l@{~}|l@{}} \toprule
Amdo Tibetan (28)&          Haitian Creole (28)  &         Polish (28)\\                   
Arabic (28) &               Hindi  (28) &                 Punjabi (28)\\                  
Bengali (28) &              Hungarian (28) &                  Romanian (28)        \\
Chinese  (28)        &       Italian (28)   &         Russian (10) \\                   
Danish (28) &               Japanese (28)   &         Spanish  (28) \\             
Dutch (8)     &              Khams Tibetan (28)&     Swedish (28)\\                 
Esperanto (28) &            Korean (28)  &            Tibetan (28)\\
German  (28) &              Kurdish  (28) &          Turkish (28)\\                 
Greek  (28) &               Pashto (28)&              Urdu (28)\\                   
Gujarati (4) &              Persian (28) &            Vietnamese (28)\\    
& &      Yue Chinese (28)\\    \bottomrule
 \end{tabular}}                                  
\caption{\label{tab:stats-per-language} Number of stories (within parentheses)
    translated from English in our ASPEN dataset.} 
\end{table}

\paragraph{Dataset Creation} Our dataset, \dataset, is based on 
the Global \textbf{A}frican \textbf{S}tory \textbf{P}roject which we
collate to serve the crosslingual g\textbf{EN}eration task sketched
above.  Global ASP  consists of over 40~stories and their
translations into~55 languages.  In creating \dataset, we only include
stories with an English version so that English plans can be created,
and (in experiments) focus on target stories in Russian, Italian, and
German. However, we are releasing stories in 31 languages together
with the associated plans (see Table~\ref{tab:stats-per-language}
  for the distribution of stories per language).

\href{https://global-asp.github.io/about/}{Global~ASP} stories are
based on illustrated children's books, and as such a few of them do
not have a narrative plot, they mostly contain short demonstrative
sentences (e.g., ``Here is a zebra. Here is an elephant.''). In
constructing \dataset, we only include stories whose average sentence
length (in the English version) is over 6~tokens.\footnote{By tokens
  here we refer to SentencePieces obtained from the mT5-large
  pretrained Tokenizer.}
Table~\ref{tab: dataset statistics} provides statistics of the
dataset. We measure
story length  in SentencePieces and the number of
sentences is calculated using a sentence splitter based on
\texttt{NLTK} \cite{bird2009natural}). Plans are created manually
based on English stories, while stories in the target languages
(\texttt{ru, de, it}) are their parallel translations. We provide
details on the content and structure of the plans in
Section~\ref{sec:methodology}.  Note that Russian comprises only 10
suitable stories. We discuss in Section~\ref{sec:Setup} how we create
training/test partitions for our few-shot experiments. The dataset is available at \hyperlink{https://github.com/evgeniiaraz/aspen_dataset/}{ASPEN URL}.

Given its small size, \dataset\ is intended primarily as an evaluation
corpus. However, it contains several languages (see
Table~\ref{tab:stats-per-language}), even though to make evaluation
tractable, we concentrate on three.


\begin{table}[t]
\centering
\footnotesize
\begin{tabular}{@{}cccc@{}}
\toprule
Languages  & Length    & Num Sents   & Num Stories \\ \midrule
  en &  482.36 & 33.61 & 28   \\
  de &  540.82 & 29.32 & 28    \\  
  ru &  490.80 & 27.80 & 10 \\
  it &  580.89 & 33.25 & 28 \\ \bottomrule
\end{tabular}%
\caption{Statistics of \dataset~ for the focus languages, German,
  Russian and Italian. Length is measured in SentencePiece tokens
  (based on mT5-large).}
\label{tab: dataset statistics}
\end{table}



\begin{figure*}[t]
  \scriptsize
  \begin{center}
    \begin{tabular}{@{}l@{~~~~}p{13cm}@{}}
      \rowcolor{Emerald!20}
\textbf{Three-act structure} & 
\emph{What is the setup?}  Goat, Dog and Cow took a taxi together. \emph{What is the main conflict?}  They needed to pay for the ride. 
\emph{What solution do the characters find?}  Cow paid her fare. Dog paid a bit extra. Goat ran away without paying
anything.
\emph{What does the story end with?}  Dow runs towards the car to find the driver who owes him change. Cow
Crosses the road calmly because she pays her fare. Goat runs away from
the car ecause she is afraid to be arrested.  \\
      \rowcolor{Emerald!20}
\textbf{Language} &  German \\
      \rowcolor{Emerald!20}
\textbf{Story} &  Ziege, Hund und Kuh waren gute Freunde. Eines Tages machten sie eine Fahrt mit einem Taxi. Als sie das Ende ihrer Fahrt erreichten, forderte der Fahrer sie auf, ihren Fahrpreis zu bezahlen. Kuh zahlte ihren Fahrpreis. Hund zahlte etwas mehr, weil er nicht das richtige Geld hatte. Der Fahrer wollte Hund gerade sein Wechselgeld geben, als Ziege weglief, ohne etwas zu bezahlen. Der Fahrer war sehr verärgert. Er fuhr davon, ohne Hund sein Wechselgeld zu geben. Deshalb rennt Hund auch heute noch zu einem Auto, um hineinzuspähen und den Fahrer zu finden, der ihm sein Wechselgeld schuldet. Ziege rennt vor dem Geräusch eines Autos davon. Sie befürchtet, verhaftet zu werden, weil sie ihren Fahrpreis nicht bezahlt hat. Und Kuh stört es nicht, wenn ein Auto kommt. Cow nimmt sich Zeit, die Straße zu überqueren, weil sie weiß, dass sie ihren Fahrpreis vollständig bezahlt hat.\\
        \rowcolor{GreenYellow!20}
\textbf{Three-act structure} &  \emph{What is the setup}
Andiswa wanted to play soccer. \emph{What is the main conflict?} The
coach told Andiswa that only boys are allowed to play
soccer. \emph{What solution does the character find?} During a big
mtch the best soccer player was ill. Andiswa asked the coach to play
instead of him. The coach allowed her to play. \emph{What does the
story end with?} Andiswa scored a goal. The crowd was joyful. Since
then the girls are allowed to play soccer.\\
        \rowcolor{GreenYellow!20}
\textbf{Language} &  German \\
        \rowcolor{GreenYellow!20}
\textbf{Story} & \\

  \end{tabular}
  
\end{center}
\vspace*{-2.5ex}
    \caption{Example prompt for crosslingual story generation.  First, a few training examples are
      demonstrated to the model (in green rectangle) and then the model
      has to complete a prefix for a test example (in yellow rectangle).} 
    \label{fig:story_prompt}
    \vspace{-2mm}
\end{figure*}

\section{Problem Formulation}
\label{sec:methodology}


We next provide an overview of our approach. We follow the few-shot
prompting line of research where a PLM is prompted with a few labelled
examples and then tasked to complete a test example based on user
input.

\subsection{Prompt Engineering}
Recent work has demonstrated the success of \emph{prompting}
general-purpose language models on a wide range of tasks, without the
need for modifying any of the model's parameters. Essentially, the
model is given an instruction in natural language describing the task
at hand (zero-shot setting), optionally followed by demonstrations,
i.e., a small number of instances exemplifying how the task is solved
(few-shot setting).

Previous work \citep{shi2022language} has shown that model output can
be fragile and highly dependent on the formulation of the prompt,
especially in multilingual settings. Following
\citet{reynolds2021prompt}, we use ``anthropomorphic'' prompts,
i.e.,~prompts which would be self-explanatory when shown to a
person. Our prompt is illustrated in Figure~\ref{fig:story_prompt}. As
can be seen, a demonstration includes the plan of a story (in
English), the name of the target language, and the target story
verbalizing the plan. It is followed by a test prefix which
generates a story based on the provided plan and target language
id. In experiments, we used three plan-story demonstrations, however,
we only show one in Figure~\ref{fig:story_prompt} for the sake of
brevity.

\begin{table}[!t]
\resizebox{0.45\textwidth}{!}{%
\begin{tabular}{@{}l@{}}
\toprule
\multicolumn{1}{c}{Story Prefixes}\\ \midrule
1. Story:                                                                          \\
2. A native \textless{}tgt\_language\textgreater~speaker would write the story as: \\
3. Die Geschichte: (\textless{}``Story:'' in tgt\_language\textgreater{}):           \\
4. \textless{}tgt\_language\textgreater~story:                                     \\ \bottomrule
\end{tabular}%
}
\vspace*{-.5ex}
\caption{Story Prefixes used within different prompts. <tgt\_language>
  is a paceholder for German, Russian, or Italian in our
  experiments. ``Die Geschichte'' is the translation of ``Story'' in
  German (for Russian it would be \foreignlanguage{russian}{историю},
  and ``Storia'' for Italian).}
\label{tab:prefixes}
\end{table}

An important variable in the prompt is the story prefix. This phrase
directly precedes the example story shown to model and is at the very
start of the instance the model is supposed to complete (at generation
time). As it directly precedes the generated story, this part of the
prompt greatly influences the output. We experimented with the four
story prefixes shown in Table~\ref{tab:prefixes}.




 \begin{figure*}[t]
   \scriptsize
   \begin{tabular}{lp{13cm}}
             \rowcolor{Orange!30}
\textbf{Story Completion} &  Goat, Dog, and Cow were great friends. One
day they went on a journey in a taxi. \\
             \rowcolor{OrangeRed!30}
\textbf{Entities} & Goat, dog, Cow, friends, they, they, the driver,
them, their ares, Cow, Dog, he, The driver, Dog, Goat, The
driver, He, Dog, Dog, the driver, him, Goat, She, she, Cow, Cow, she,
she \\
             \rowcolor{Magenta!30}
\textbf{Plot Outline} &  Goat, Dog and Cow are good friends. They take a trip together on a taxi, but when the time comes to pay the
driver one of the friends does something surprising. \\
             \rowcolor{Lavender!30}
\textbf{Three-act Structure} & \emph{What is the setup?} Goat, Dog and Cow took a taxi together. \emph{What is the main conflict?} They needed to pay for
the ride. \emph{What solution do the characters find?} Cow paid her fare. Dog paid a bit extra. Goat ran away without paying anything. \emph{What does the story
end with?} Dog runs towards the car to find the driver who owes him change. Cow crosses the road calmly because she paid her fare. Goat runs away
from the car because she is afraid to be arrested. \\
\end{tabular}
    \caption{Story plan examples. Questions in italics correspond to
      main events in three-act structure.}
    \label{fig:story_plans}
    \vspace{-2mm}
\end{figure*}


 \subsection{Plan Representation}
\label{sec:Plans}
In our crosslingual setting, we rely on plans to convey the content of
the story to the model.  In experiments, we use manually created plans
which vary in terms of their form and the level of plot detail
supplied to the model. We describe these below and provide examples in
Figure~\ref{fig:story_plans}.

\paragraph{Story Completion} Much previous work 
has focused on generating a continuation or ending for an incomplete
story
\cite{mostafazadeh-etal-2016-corpus,fan2018hierarchical,ijcai2019p727,mori-etal-2022-plug}.
We adapted story completion to our crosslingual setting, by prompting
the model with the first two sentences of the English story. Although
the beginning of the story is not a plan as such, we assume it
contains enough detail about the characters and the setting of the
story (see Figure~\ref{fig:story_plans}). Our intuition is that the
model will learn that the task involves translating the incomplete
story to the target language and then generating a continuation.

\paragraph{Entities} Content 
 planning strategies based on entities have been proven effective in a
 variety of tasks beyond story generation, including summarization
 \cite{narayan-etal-2021-planning,liu-chen-2021-controllable} and
 data-to-text generation \cite{puduppully-lapata-2021-data}. Entities
 also play a pivotal role in various theories of discourse which posit
 that coherence is achieved in view of the way discourse entities are
 introduced and discussed \cite{Grosz:ea:95}. Our entity-based plans
 are a list of characters, places, and objects in the story (see
 Figure~\ref{fig:story_plans}).

 \paragraph{Plot Outline} Much research on story generation \citep[][\emph{inter
   alia}]{goldfarb2020content,rashkin-etal-2020-plotmachines,fan2019strategies}
 has resorted to plot outlines as a means of instilling generation
 models with knowledge about events and their logical
 progression. Stories in  our dataset are accompanied by short,
 high-level summaries which give an overview of the plot without going
 into detail.  We used these summaries as a proxy for plot outlines
 (see the example in Figure~\ref{fig:story_plans}).

\paragraph{Three-act Structure} The three-act structure is a model
used in narrative fiction that divides a story into three parts
(acts), often called the Setup, the Confrontation, and the Resolution.
\citep{McKee:1997,field2005screenplay}. The setup describes the
overall circumstances of the story, e.g.,~main characters and their
life. The confrontation describes the conflict which the main
character needs to overcome. Resolution describes the steps the
protagonist takes to resolve the conflict and their outcome. We
devised plans based on the three-act structure under the hypothesis
that stories attempt to answer the following questions: What is the
setup?, b)~What is the main conflict?, c)~What solution does the
character find?, and d)~What does the story end with? All plans have
the same questions, however, the answers vary depending on the content
of individual stories (see Figures~\ref{fig:story_prompt}
and~\ref{fig:story_plans}).

\section{Experimental Setup}
\label{sec:Setup}

\subsection{Models} 

As mentioned earlier, we employ a single model for all story
generation experiments. Specifically, we use PaLM
\cite{chowdhery2022palm} \footnote{https://developers.generativeai.google/}
which was mostly trained on English (80\% of training data). However,
PaLMs have recently demonstrated impressive results in multilingual
settings, especially when prompted with several training examples
\cite{winata-etal-2021-language,chowdhery2022palm, shi2022language}.
Throughout this paper, PaLM was called with temperature $\tau=0.7$,
allowing for some randomness in the output, with a beam width
of~$w=10$. All experiments used the prompts described in
Section~\ref{sec:methodology}, with three (randomly sampled)
plan-story examples per language. We use the same three examples for
all languages and treat the remaining stories in \dataset~ as test
data. Plans varied along the dimensions introduced in
Section~\ref{sec:Plans}.

As a baseline, we used another strong state-of-the-art {multilingual} model, namely
mT5 \cite{xue2021mt5}. mT5 was pretrained on data in
multiple languages, making it especially suited to our crosslingual
generation task. We finetuned mT5-XL \cite{xue-etal-2021-mt5} on three plan-story examples per language\footnote{The prompt text was not included in mT5 finetuning, only the plans.} (same as those seen by the PLM) for 500 steps with a batch size of 8 and a learning rate of 0.0001. We selected the last checkpoint for each language. 
As an upper bound we further
translated the English stories into the target language using the
public Google Translate API \cite{Wu:Ea:2016}.

\subsection{Evaluation}

Story generation is a creative task, the model is
allowed to generate new content which is not included in the input
\citep{dengetal2021compression}. In our case, not only is the model
creating new text, but also the story is expected to be in a different
target language. These constraints make evaluation challenging, the
generated text needs to be fluent in the target language (no
repetitions/grammar errors), while the story needs to be overall coherent,
and faithful to the English plan. We evaluate these aspects both
automatically and in a judgment elicitation study.

\begin{table*}[t]
\centering
\footnotesize
\resizebox{\textwidth}{!}{%
\begin{tabular}{l@{~~}l|ccc|ccc|ccc|ccc}\toprule
 & &\multicolumn{3}{c}{DE} &\multicolumn{3}{c}{RU} &\multicolumn{3}{c}{IT} &\multicolumn{3}{c}{AVG} \\
& \multicolumn{1}{c|}{Models} &VocTok $\uparrow$ & Inter $\downarrow$
 &Intra  $\downarrow$ &VocTok  $\uparrow$  &Inter  $\downarrow$ &Intra
$\downarrow$ &VocTok $\uparrow$&Inter $\downarrow$ &Intra $\downarrow$ &VocTok   $\uparrow$ &Inter $\downarrow$ &Intra $\downarrow$ \\ \hline
 & Entities  &0.39 &22.63 &1.75 &0.34 &48.36 &3.15 &0.37 &26.62 &1.35 &0.37 &32.53 &\textbf{2.09} \\ 
&  Story Completion  &0.43 &33.53 &4.04 &0.38 &51.08 &\textbf{0.72} &0.61 &23.09 &3.81 &0.47 &35.90 &2.86 \\
&  Plot Outline  &0.41 &38.06 &10.03 &\textbf{0.65} &22.09 &2.64 &0.47 &21.18 &\textbf{1.03} &0.51 &27.11 &4.57 \\
\raisebox{.2cm}[0pt]{\begin{sideways}PaLM\end{sideways}} & 3Act
Structure  &\textbf{0.57} &\textbf{18.85} &\textbf{1.68} & 0.58 &\textbf{14.66} &3.51 &\textbf{0.48} &\textbf{16.00}
&1.52 &\textbf{0.54} &\textbf{16.50} &2.24 \\\hline
& mT5 & 0.58 & 91.17 & 0.00 & 0.73 & 68.34 & 0.42 & 0.68 & 87.77 & 0.00 & 0.66 & 82.43 & 0.14\\
& Google Translate & 0.57 & 6.85 & 0.74 & 0.70 & 4.08 & 0.21 & 0.58 & 7.23 & 0.42 & 0.62 & 6.05 & 0.46 \\ 
& Reference & 0.58 & 8.02 & 0.50 & 0.65 & 5.16 & 2.22 & 0.57 & 7.61 & 0.52 & 0.60 & 6.93 & 1.08 \\
\bottomrule
\end{tabular}
}
\caption{Diversity and repetitiveness metrics for PaLM  with
  different plan variants and comparison models.}
\label{tab:div}
\end{table*}

\paragraph{Automatic Evaluation} 


We evaluated various aspects of fluency following the automatic
metrics introduced in \citet{goldfarb2020content}. Specifically, we
use \emph{vocabulary-to-token ratio} to measure the extent to which
the vocabulary of the generated stories is repetitive (higher is
better).  We further measure \emph{intra-story repetition} as a
fluency metric using the proportion of trigrams which are repeated
\emph{within} a story (lower is better). Finally, we also compute
\emph{inter-story repetition} as a diversity metric to quantify
whether the model has learnt to generate only one ``kind'' of text
irrespective of the input. We measure the proportion of trigrams
repeated \emph{between} stories (lower is better). 

We also use MAUVE \cite{pillutla2021mauve} to measure the naturalness
of the generated stories.  MAUVE is a recently introduced automatic
metric for open-ended generation which has high correlation with human
judgements. It computes the similarity of the distribution of
human-written text and machine-generated text, while being sensitive
to to generation length, different decoding algorithms, and model
size. Machine-generated distributions in MAUVE are computed with GPT-2
\cite{radford2019language}. We use gold reference stories in the
target language as our set of human written texts.



Finally, we also evaluate the similarity of model output against
reference stories in \dataset. As the stories are parallel in the
dataset, we expect that the closer the model follows the plan based on the 
English story, the more similar the resulting story will be to the
golden stories. Notably, the reference stories are manual native spaker translations of the stories to the target languages. We chose these as golden stories to compare against non-creative, translation only baseline. We use SentencePieceROUGE \cite{vu2022sprouge}, a
ROUGE-inspired \cite{lin-2004-rouge} metric which uses
language-independent SentencePiece
\cite{kudo-richardson-2018-sentencepiece} tokenization.

\paragraph{Human Evaluation}
We also carry out human evaluation following a simplified version of
the annotation protocol proposed in
\citet{chhun-etal-2022-human}. Specifically, crowdworkers are
presented with the English prompt given to the model,
the human-authored story in the target language, followed by the
machine generated story, also in the target language. We provided
reference stories so that participants could better calibrate their judgment
\cite{karpinska-etal-2021-perils}.

Participants are asked to rate the stories along the following
dimensions: (a) \emph{Relevance} measures whether the story matches
its prompt; (b) \emph{Fluency} measures the quality of the text
including grammatical errors and repetitions; (c) \emph{Coherence}
measures whether the story's plot makes logical sense; and (d)
\emph{Engagement} measures the extent to which the reader engaged with
the story.  Each story was evaluated by three workers on the four
criteria using a 3-point Likert scale where 1~is worst and 3~is
best. Annotators were native speakers of the target language, and
fluent in English. We evaluate the output of our model with the three
plan configurations, mT5, and Google Translate (without a
plan). Overall, we elicited ratings for 285~stories (35 in Russian,
125 in Italian, and 125 in German). We collect ratings from three
different annotators for each data
point. Figure~\ref{fig:human_eval_instructions} in
Appendix~\ref{appendix:human:eval} presents our detailed instructions.

%
%
%


\begin{table}[t]\centering
\footnotesize
\resizebox{0.45\textwidth}{!}{%
\begin{tabular}{l@{~~}l|rrr|r}\toprule
&\multicolumn{1}{c|}{Models} &DE &RU &IT &AVG \\ \hline
& Entities &0.95 &0.91 &0.99 &\textbf{0.95} \\ 
& Story Completion  &\textbf{0.99} &0.66 &0.99 &0.88 \\  
& Plot Outline  &0.98 &0.20 &0.99 &0.72 \\  
\raisebox{.2cm}[0pt]{\begin{sideways}PaLM\end{sideways}} & 3Act Structure  &\textbf{0.99} &0.66 &0.99 &0.89 \\ \hline
& mT5  & 0.37 & \textbf{0.99} & 0.86 & 0.74\\
& Google Translate & 0.99 & 0.20 & \textbf{1.00} & 0.73 \\ 
\bottomrule
\end{tabular}
}
\caption{MAUVE for PaLM with different plan variants and comparison
  models.}
  \label{tab:mauve}
\end{table}

\section{Results}

\subsection{Automatic Evaluation}
Tables~\ref{tab:div}--\ref{tab:spROUGE} summarize our results using
automatic evaluation metrics. We present results with different
instantiations of our model according to the plans presented in
Section~\ref{sec:Plans} and the prefix
``\textless{}tgt\_language\textgreater~story:'' (see
Table~\ref{tab:prefixes}) which performed overall best. We describe
results with alternative story prefixes in
Appendix~\ref{story:prefix:full}. We also compare to an mT5 model
fine-tuned in a similar few-shot setting, and Google Translate as an
upper bound. Wherever possible we also report metrics on the gold
reference translations.

\paragraph{How Diverse are the  Generated Stories?}

In Table~\ref{tab:div} we report results with vocabulary to token
ratio (VocTok), Inter- and Intra-story repetition for each target
language, and on average. As can be seen, stories following the
three-act structure have the most diverse vocabulary (see VocToK
column), they also tend to be overall fluent (see Intra column).
Inter-story repetition measures whether the constructions used are
repetitive across stories.  Ideally, we would like to avoid cases
where the model generates the same story no matter the plan.
Table~\ref{tab:div} shows that the three-act plan leads to more
diverse texts. We hypothesize it forces the model to focus on main
events and their ordering while other plan formulations exercise less
control on the content and structure of the story. The mT5 model performs poorly, especially with respect to the Inter-story repetition metric, which indicates that it generates the same story irrespective of the plan. The Google Translate upper bound performs quite well, approaching  human parity (see row Reference in Table~\ref{tab:div}).

\paragraph{Do Stories Resemble Human Writing?}

\begin{table}[t]\centering
\footnotesize
\resizebox{0.45\textwidth}{!}{%
\begin{tabular}{l@{~~}l|r|rrr|r}\toprule
& \multicolumn{1}{c|}{Models} &EN &DE &RU &IT &AVG \\ \hline
& Entities  & 21.52 &20.19 &20.32 &19.92 &20.14 \\ 
& Story Completion  & 20.07 &19.71 &14.17 &20.24 &18.04 \\  
& Plot Outline  & 19.38 &18.10 &18.46 &18.60 &18.39 \\  
\raisebox{.2cm}[0pt]{\begin{sideways}PaLM\end{sideways}} & 3Act Structure  & 23.52 &\textbf{22.61} &\textbf{21.16} &\textbf{24.34} &\textbf{22.70} \\\hline
& mT5 & N/A & 15.84 & 12.70 & 15.25 & 14.60 \\
& Google Translate & N/A & 73.98 & 68.31 & 68.88 & 70.39\\
\bottomrule
\end{tabular}
}
\caption{SentencePiece-ROUGE between generated and reference  story
  for PaLM with different plan variants and comparison models. EN results are provided for reference and excluded from AVG.}
\label{tab:spROUGE}
\end{table} 

We now examine whether the generated stories are similar to
human-written texts. Table~\ref{tab:mauve} quantifies the naturalness
of the stories using MAUVE \cite{pillutla2021mauve}. In general, we
observe that the similarity of model output against human-written text
is high, particularly for German and Italian. Plans based on
entities impose least constraints on the output, and therefore lead to most natural text. Plans following
the three-act structure are second best according to
MAUVE. We also see that MAUVE penalizes Google Translate, we suspect it is prone to translationese which renders the stories less natural. The same is true for the mT5 model. 

\paragraph{Are Machine and Reference Stories Similar?}

In Table~\ref{tab:spROUGE}
we evaluate similarity between automatically generated
and ``gold'' reference stories using SentencePiece-ROUGE. This allows
us to  simultaneously measure  whether the events in the gold story are mentioned
in the generated story and whether the language usage is similar
between reference stories and automatically generated ones. 
Across all languages, we observe that stories based on the three-act
structure are more similar to the reference. This is not surprising
given this plan is most detailed; the story completion and plot
outline plans cover most of the content of the story but lack
details. The model based on entity plans  performs quite well even though the plan is not very elaborate.  These stories tend to be fairly long (see Table~\ref{tab: stories lengths}) and thus favored by  recall-oriented ROUGE. mT5 struggles to generate stories which follow the plan is penalized by ROUGE. In general, all models have a long way to go compared to the upper bound (see Google Translate in the table). 

\begin{table}[t]
\small
\resizebox{\linewidth}{!}{%
\begin{tabular}{@{}l@{~~}l|crrll@{}}
\toprule
& Models        & \multicolumn{1}{c}{DE}      & \multicolumn{1}{c}{RU}      & \multicolumn{1}{c}{IT}      & \multicolumn{1}{c}{AVG} \\ \midrule
& Entities    & 551.50   & 136.39 & 656.15 &  448.01   \\
& Story Completion  & 505.50   & \textbf{94.23}  & 453.15 &  350.96  \\
& Plot Outline     & 529.52 & 124.65 & 541.27 &   398.30  \\
\raisebox{.2cm}[0pt]{\begin{sideways}PaLM\end{sideways}} &3Act Structure & \textbf{386.62} & 99.81  & 431.31 &    \textbf{305.91} \\\hline
&mT5 & 494.96 & 260.00 & \textbf{355.08} & 370.01 \\
&Google Translate & 557.08 & 461.71 & 598.12 & 538.97\\
&Reference & 548.08 & 453.29 & 585.32  & 528.90 \\
\bottomrule
\end{tabular}%
}
\caption{Average story length (measured in SentencePiece tokens) for PaLM
 and comparison models.}
\label{tab: stories lengths}
\end{table}


\begin{table}[t]
\centering
\footnotesize
\begin{tabular}{@{}l@{~~}lc@{}}
\toprule
& \multicolumn{1}{c}{Models} & AVG         \\ \hline
& Entities & 66.66   \\
& Story Completion & 42.31 \\
& Plot Outline      & 59.64\\ 
\raisebox{.2cm}[0pt]{\begin{sideways}PaLM\end{sideways}} & 3Act Structure &   38.58  \\\hline
& mT5 & 77.19 \\
& Google Translate & 66.66 \\
& Reference & 68.42 \\
\bottomrule
\end{tabular}%
\caption{Percentage of stories containing direct speech for PaLM and
  comparison models.}
\label{tab: percentage direct speech}

\end{table}

\paragraph{Do Stories Have Different Length?}
\label{subsec: length}

The plans can be viewed as constraints on open-ended generation. In
other words, we expect that with plans containing more information,
the model will have a clearer representation of the content of the
story and will stop generating once all of the planned content is
covered. To test this hypothesis, we compare the average length of
stories generated based on different plans (we measure length in
SentencePieces based on the vocabulary pretrained for mT5-large). The results in Table~\ref{tab: stories lengths}
corroborate our hypothesis. Plans based on three-acts contain most
information about events in the story and their order, resulting in
the shortest narratives. In contrast, entity-based plans are
least constraining, they do not contain information about relations
between characters or the order of events and as a result the model
has free reign to fill in missing content and ends up generating longer stories. All models, including mT5, generate shorter stories compared to the Reference and Google Translate.

\paragraph{Do Stories Contain Direct Speech?}
The plans used in our experiments are prosaic, containing no examples
of direct speech. However, dialogue is a common device for rendering
narratives more engaging, moving the story forward, and allowing
characters to engage in conflict \cite{Scott-Bell:2014}. We evaluate the extent to which
a model is creative by computing the percentage of stories which
include direct speech. We assume direct speech is marked with pairs of
quotation marks.  The results in Table~\ref{tab: percentage direct
  speech} show that plans based on entities and three-acts display the
most and least direct speech, respectively. This further confirms our
observation that with less constraining plans the model is being more
creative with its output, but  at the cost of  coherence (as we
discuss below). Interestingly, the percentage of stories with direct speech generated by mT5 is seemingly high; this is because  the model generates the same story irrespective of the input plan in most cases. As this story contains an example of direct speech, this percentage is high.  




\begin{table}[t]
\centering
\footnotesize
\resizebox{0.5\textwidth}{!}{%
\begin{tabular}{@{}l@{~~}l|cccc@{}}\toprule
& \multicolumn{1}{c|}{Models} & Relevance & Fluency & Coherence & Engagement \\ \hline
& Entities  & 0.37$^\diamond$ & 0.51$^\diamond$$^{\;\;}$ & 0.46$^\diamond$ & 0.33$^\diamond$ \\ 
&  Plot Outline & 0.36$^\diamond$  & 0.55$^\diamond$$^{\;\;}$ & 0.50$^\diamond$ & 0.31$^\diamond$ \\
\raisebox{0cm}[0pt]{\begin{sideways}PaLM\end{sideways}} & 3Act
Structure  & \textbf{0.59}$^{\;\;}$  & 0.61$^\diamond$$^\ast$ & \textbf{0.69}$^\ast$ & \textbf{0.46}$^\diamond$ \\
& mT5 & 0.00$^{\;\;}$ & \textbf{0.82}$^{\;\;}$$^{\;\;}$ & \textbf{0.69}$^\ast$ & 0.42$^\diamond$ \\\hline
& Google Translate & 0.99$^{\;\;}$ & 0.68$^\ast$$^{\;\;}$ & 0.92$^{\;\;}$ & 0.60$^{\;\;}$\\ 
\bottomrule
\end{tabular}
}
\caption{Average human evaluation results across languages for PLM  with different plan variants and comparison models. Best results in upper block are \textbf{boldfaced}. Systems in each column are marked
  with same symbols when differences between them are not
  statistically significant; unmarked pairwise differences are
  significant (using a one-way ANOVA with post-hoc Tukey HSD tests;
  \mbox{$p < 0.01$)}.}
\label{tab:heval-average}
\end{table}

\subsection{Human Evaluation}
\label{sec:human-evaluation}

Our judgment elicitation study
is summarized in Table~\ref{tab:heval-average}. 
Our results are averaged across
languages, we present ratings for each language in  Appendix~\ref{humaneval:full}). Human evaluation confirms our automatic results. The model based on the three-act structure manages to generate stories which `stick' to the plan  (see Relevance column), whereas models based on entities and plot outlines are less faithful (the difference between the these two and the three acts is statistically significant), while mT5 largely ignores the plan.  All PLM variants are equally fluent (differences among them are not statistically significant). The three-acts model is as fluent as Google Translate, however, the most 
fluent model is mT5 which is good at generating text but not necessarily the story we want. 

Human raters find the stories based on the three-acts as coherent as
those obtained from mT5 and both of these (significantly) less
coherent than Google Translate.  The stories obtained from the three
acts are more engaging compared to entities and plot outlines, but
less so compared to the Google Translate upper bound (which is
significantly better against all other models).  We show examples of
stories our raters saw in Table~\ref{tab: sample stories1}. We present
 analysis of system output in
Appendix~\ref{sec:qual-eval}.

\begin{table}[t]
\scriptsize
\begin{tabular}{p{7cm}} \toprule
  \multicolumn{1}{c}{Entities} \\ \midrule
Una volta, c'era \colorbox{lime}{una mucca, un cane e una capra} che andarono a visitare un altro villaggio. Decisero di prendere un autobus. Il cane era un po' preoccupato, ma la capra era molto felice e disse al cane: "Non preoccuparti! La gente sarà gentile con noi." Il cane non era sicuro, ma la capra era determinata. Così tutti e tre i loro amici salirono sull'autobus. Quando \colorbox{lime}{il conducente} dell'autobus li vide salire, cominciò a ridere. "Siete voi tre \colorbox{lime}{amici} che vi siete persi?" Disse  \colorbox{lime}{il conducente}. "Dove state andando?" \colorbox{lime}{La capra}, che era più intelligente dei \colorbox{lime}{suoi amici}, rispose: "Stiamo andando a visitare un altro villaggio." \colorbox{lime}{Il conducente} dell'autobus si mise a ridere di nuovo. "Bene, amici! Non ho mai visto animali sull'autobus! Siete molto divertenti!" Disse.\colorbox{lime}{Il cane} non era sicuro. \colorbox{lime}{La capra} rispose: "Abbiamo pagato abbastanza soldi per prendere l'autobus."\\ \toprule
\multicolumn{1}{c}{3Act Structure}  \\ \midrule
\colorbox{lime}{Una volta, la mucca, la capra e il cane viaggiarono insieme sul taxi.}  <...> Chiese la mucca. "Se non
pagherai, il conducente del taxi chiamarà la polizia e sarai
arrestata.Non puoi semplicemente scappare!" \colorbox{lime}{La capra
  non ascoltò e corse via. Il cane si precipitò verso
  il taxi per} \colorbox{lime}{chiedere il resto all'autista. La mucca
  attraversò la strada con calma,} \colorbox{lime}{perché aveva pagato
  il giusto prezzo. La capra uscì correndo dal taxi} \colorbox{lime}{e scappò via.} La capra non pagò il
taxi. La capra fuggì. \colorbox{lime}{La mucca attraversò la strada con calma.}\\ \bottomrule
\end{tabular}%
\caption{Abridged stories generated in Italian from different plans (story ID 4: ``Goat, Dog and Cow''). 
Highlighted excerpts correspond to the events and/or characters
mentioned in the input plan.}
\label{tab: sample stories1}
\vspace{-0.1cm}
\end{table}

\section{Conclusions}

In this work we considered the problem of automatically generating
stories in multiple languages by prompting pretrained language models with
different plans. We investigated which plan formulation is better
suited to our crosslingual generation task and empirically
demonstrated that plans following the three-act structure generate
more coherent narratives. To facilitate research in this area, we
further collated \dataset, a new dataset with multilingual stories and
plan annotations.

In the future, we would like to explore further the
potential of formulating plans as question-answer pairs, to generate
stories that are both creative and controllable. For example, we could
generate more detailed questions with very short answers or no answers
at all. It would also be interesting to generate the plan and the
story iteratively allowing human writers to intervene, and directly
steer the generation process. Further, our experiments have not explored how the number of example stories provided to the model at inference time influences the output in a creative task such as story generation. We leave it as a potential future research. 

\noindent In this paper we mainly focused on three Indo-European languages, namely
 Russian, German, and Italian. Although the proposed methods are in principle language agnostic, the PLM's performance might vary
 depending on the number and amount of languages it has been trained on. Another limitation concerns the use of the large language model itself which  is feasible only with large computational resources. Benchmarking LLMs on ASPEN is out of scope of this paper and we leave it as future work, together with other cross-lingual transfer methods, e.g., translation-based. Finally, throughout this paper, we have assumed that the plans are provided to the PLM, e.g., in an interactive setting where a user thinks of a story sketch, and the system fleshes out the details. In the future, it would be interesting to consider generating plan candidates as well as stories.

\bibliographystyle{lrec-coling2024-natbib}
\bibliography{lrec-coling2024-example}

\clearpage

\appendix




\section{Human Evaluation}
\label{appendix:human:eval}

Figure~\ref{fig:human_eval_instructions} presents the instructions shown to crowdworkers during our human evaluation study. To recruit our participants, we screened their language skills to determine whether they're native speakers, their education level and country of residence as well as origin. There were a total of 18 raters, of which 38.89\% are from Germany, 38.89\% from Ukraine and the rest were from Italy. 61.11\% of them holds a master degree, 27.78\% of them holds a High school degree or equivalent and 11.11\% holds a Bachelor's degree. All our annotators are paid adequately by our suppliers adhering to the supplier code of conduct. 

A detailed break down of our human results is provided in Table~\ref{tab:heval}.

\label{humaneval:full}
\begin{figure*}[t]
    \centering
    \footnotesize
    \resizebox{0.95\textwidth}{!}{
    \begin{tabular}{@{}p{15cm}@{}} 
    \hline
    \textbf{Task:} Please read the prompt, the human-written story and the machine-generated story, carefully. We will then ask you to rate the machine-generated story. \\
    The prompt will always be in English, but the stories will be in one of the following target languages: German, Italian, or Russian. Both the human-written story and the machine-generated story will be presented in the same target language. \\ \hline
    \hline
    \textbf{Prompt:} <List of Entities, or List of question-answer pairs or a small summary> in English.   \\ \hline
    \hline
    \textbf{Human-written Story:} <Human written story in a target language> \\ \hline
    \hline
    \textbf{Machine-generated Story:} <Machine-generated story in the same target language>     \\ \hline
    \hline
    \textbf{Q1: [Relevance -- measures how well the story matches its prompt]} \\
    \textbf{1} -- The story has no relationship with the prompt at all (it has a lot of extra details not mentioned in the prompt). \\
    \textbf{2} -- The story roughly matches the prompt (but has a few extra details not mentioned in the prompt). \\
    \textbf{3} -- The story matches the prompt exactly.\\ \hline
    \hline   
    \textbf{Q2: [Fluency -- measures the quality of the text. Please make sure not to evaluate the plot of the story, but just the language.]} \\
    \textbf{1} -- Hard to understand with multiple grammatical errors and/or multiple repetitions.  \\
    \textbf{2} -- Possible to understand but there are some mistakes/repetitions. \\
    \textbf{3} -- Easy to understand without any mistakes/disfluencies.\\ \hline
    \hline   
    \textbf{Q3: [Coherence -- measures whether the plot of the story makes sense]} \\
    \textbf{1} -- The story does not make sense at all. For instance, the setting and/or characters keep changing, and/or there is no understandable plot. \\
    \textbf{2} -- The story mostly makes sense but is incoherent at places. \\
    \textbf{3} -- The story makes sense overall.\\ \hline
    \hline   
    \textbf{Q4: [Engagement -- measures how much you engaged with the story] } \\
    \textbf{1} -- You mostly found the story boring. There may be one or two things interesting in the story, but no more. \\
    \textbf{2} -- The story was mildly interesting. \\
    \textbf{3} -- The story almost kept you engaged until the end. \\ \hline
    \end{tabular}
    } 
    \caption{Instructions used in our  human evaluation.}
    \label{fig:human_eval_instructions} 
\end{figure*}

\begin{table*}[t]
\centering
\footnotesize
\resizebox{\textwidth}{!}{%
\begin{tabular}{l@{~~}l|cccc|cccc|cccc}\toprule
 & &\multicolumn{4}{c}{DE} &\multicolumn{4}{c}{RU} &\multicolumn{4}{c}{IT} \\
& \multicolumn{1}{c|}{Models} & Relevant & Fluent & Coherent & Engaging & Relevant & Fluent & Coherent & Engaging & Relevant & Fluent & Coherent & Engaging \\ \hline
& Entities  & 0.38 & 0.58 & 0.51 & \textbf{0.38} & 0.24 & 0.45 & 0.45 & 0.33 & 0.49 & 0.51 & 0.40 & 0.27\\ 
&  Plot Outline & 0.34 & 0.53 & 0.61 & 0.28 & 0.21 & 0.60 & 0.36 & 0.33 & 0.51 & 0.53 & 0.54 & 0.31 \\
\raisebox{.2cm}[0pt]{\begin{sideways}PaLM\end{sideways}} & 3Act
Structure  & \textbf{0.49} & 0.66 & 0.73 & 0.37 & \textbf{0.57} & \textbf{0.64} & \textbf{0.69} & \textbf{0.62} & \textbf{0.69} & 0.51 & \textbf{0.64} & \textbf{0.39} \\
& mT5 & 0.00 & \textbf{0.87} & \textbf{0.86} & 0.37 & 0.00 & 0.81 & 0.67 & 0.52 & 0.00 & \textbf{0.79} & 0.55 & 0.37 \\\hline
& Google Translate & 0.97 & 0.83 & 0.94 & 0.51 & 1.00 & 0.69 & 0.93 & 0.69 & 0.99 & 0.51 & 0.89 & 0.59\\ 
\bottomrule
\end{tabular}
}
\caption{Human evaluation (per language) results  for PaLM  with different plan variants and comparison models. Best results in upper block are \textbf{boldfaced}.}
\label{tab:heval}
\end{table*}

\section{Results with Alternative Story Prefixes}
\label{story:prefix:full}

Table~\ref{tab:div:full}--\ref{tab:spROUGE:full} present results for
our model when using different story prefixes in the prompt.

\begin{table*}[t]
\centering
\footnotesize
\resizebox{\textwidth}{!}{%
\begin{tabular}{lr|crr|ccr|ccc|ccc}\toprule
&   &  \multicolumn{3}{c}{DE} &\multicolumn{3}{c}{RU} &\multicolumn{3}{c}{IT} &\multicolumn{3}{c}{AVG} \\
\multicolumn{1}{c}{Plan} & \multicolumn{1}{c|}{Prefix}  &VocTok
$\uparrow$ &\multicolumn{1}{c}{Inter $\downarrow$}
&\multicolumn{1}{c|}{Intra  $\downarrow$} &VocTok  $\uparrow$  &Inter
$\downarrow$ & \multicolumn{1}{c|}{Intra $\downarrow$} &VocTok &Inter $\downarrow$ &Intra $\downarrow$ &VocTok  $\uparrow$ &Inter $\downarrow$ &Intra $\downarrow$ \\ \hline
Entities &None &0.38 &34.46 &3.06 &0.38 &33.49 &5.98 &0.50 &33.40 &2.71 &0.42 &33.78 &3.92 \\
Entities &Long &0.46 &22.47 &1.55 &0.36 &30.07 &2.77 &0.39 &26.86 &1.37 &0.40 &26.47 &1.90 \\
Entities &Target &0.43 &24.02 &1.10 &0.49 &13.30 &0.28 &0.37 &25.19 &0.74 &0.43 &20.84 &0.71 \\
Entities &Lang &0.39 &22.63 &1.75 &0.34 &48.34 &3.15 &0.37 &26.62 &1.35 &0.37 &32.53 &2.09 \\ \midrule
Story Completion &None &0.68 &8.68 &2.09 &0.60 &32.92 &8.42 &0.79 &24.8 &4.91 &0.69 &22.13 &5.14 \\
Story Completion &Long &0.38 &31.37 &4.28 &0.63 &45.06 &16.36 &0.55
&18.56 &0.62 &0.52 &31.66 &7.08 \\
Story Completion &Target &0.49 &19.33 &2.96 &0.55 &51.85 &11.62 &0.44 &30.57 &3.12 &0.49 &33.92 &5.90 \\
Story Completion &Lang &0.43 &33.53 &4.04 &0.38 &51.08 &0.72 &0.61 &23.09 &3.81 &0.47 &35.89 &2.86 \\ \midrule
Plot Outline &None &0.64 &5.22 &0.71 &0.56 &33.27 &6.99 &0.68 &21.45 &1.81 &0.63 &19.98 &3.17 \\
Plot Outline &Long &0.42 &27.88 &3.08 &0.49 &12.49 &2.72 &0.45 &29.52 &1.75 &0.45 &23.30 &2.52 \\
Plot Outline &Target &0.36 &36.60 &1.57 &0.45 &21.55 &0.79 &0.42 &22.86 &4.07 &0.41 &27.00 &2.15 \\
Plot Outline &Lang &0.41 &38.06 &10.03 &0.65 &22.09 &2.64 &0.47 &21.18 &1.03 &0.51 &27.1 &4.57 \\ \midrule
3Act Structure &None &0.49 &20.18 &0.69 &0.43 &25.86 &2.14 &0.51 &13.03 &1.44 &0.48 &19.69 &1.42 \\
3Act Structure &Long &0.57 &19.07 &1.45 &0.47 &18.70 &1.45 &0.53 &15.06 &0.97 &0.53 &17.61 &1.29 \\
3Act Structure &Target &0.46 &17.71 &1.19 &0.55 &31.14 &0.65 &0.49 &15.04 &0.96 &0.50 &21.23 &0.93 \\
3Act Structure &Lang &0.57 &18.85 &1.68 &0.58 &14.66 &3.51 &0.48 &16.00 &1.52 &0.54 &16.50 &2.24 \\
\bottomrule
\end{tabular}
}
\caption{Diversity and repetitiveness metrics for PaLM with different
  plans and story prefixes. None is a shorthand for the prefix
  ``Story:'', Long abbreviates prefix ``A native
  \textless{}tgt\_language\textgreater~speaker would write the story
  as:'', Target is the translation of the prefix ``Story:'' in the
  target language, and Lang refers to
  ``\textless{}tgt\_language\textgreater~story:''.  }
\label{tab:div:full}
\end{table*}

\begin{table}[t]\centering
\footnotesize
\resizebox{0.45\textwidth}{!}{%
\begin{tabular}{ll|lcl|c}\toprule
Plan & Prefix &\multicolumn{1}{c}{DE} &RU &\multicolumn{1}{c|}{IT} &AVG \\ \hline
Entities &None &0 &0.66 &0.99 &0.55 \\
Entities &Long &0.99 &0.20 &1 &0.73 \\
Entities &Target &0.96 &0.66 &1 &0.87 \\
Entities & Lang & 0.95 & 0.91 &0.99 &0.95 \\ 
Story Completion &None &0.74 &0.97 &0.60 &0.77 \\
Story Completion &Long &1 &0.20 &0.99 &0.73 \\
Story Completion &Target &1 &0.91 &0.77 &0.89 \\
 Story Completion & Lang &0.99 &0.66 &0.99 &0.88 \\  
Plot Outline&None &0.95 &0.99 &0.24 &0.72 \\
Plot Outline &Long &0.99 &0.91 &1 &0.97 \\
Plot Outline &Target &1 &0.86 &1 &0.95 \\
 Plot Outline &  Lang & 0.98 &0.20 &0.99 &0.72 \\  
3Act Structure &None &1 &0.99 &1 &0.99 \\
3Act Structure &Long &0.89&0.91 &0.99 &0.93 \\
3Act Structure &Target &0.99 &0.66 &1 &0.88 \\
3Act Structure & Lang  &0.99 &0.66 &0.99 &0.89 \\ 
\bottomrule
\end{tabular}
}
\caption{MAUVE for PaLM with different plan variants and story
  prefixes. None is a shorthand for the prefix
  ``Story:'', Long abbreviates prefix ``A native
  \textless{}tgt\_language\textgreater~speaker would write the story
  as:'', Target is the translation of the prefix ``Story:'' in the
  target language, and Lang refers to
  ``\textless{}tgt\_language\textgreater~story:''.}
  \label{tab:mauve:full}
\end{table}

\begin{table}[t]\centering
\footnotesize
\resizebox{0.45\textwidth}{!}{%
\begin{tabular}{ll|ccr|r}\toprule
Plan & Prefix &DE &RU &\multicolumn{1}{c|}{IT} &AVG \\ \hline
Entities & None &13.21 &12.18 &13.41 &12.93 \\
Entities & Long  &20.12 &18.87 &23.20 &20.73 \\
Entities & Target &20.44 &17.16 &21.75 &19.78 \\
Entities & Lang &20.19 &20.32 &19.92 &20.14 \\ 
Story Completion &None &10.25 &17.96 &7.96 &12.05 \\
Story Completion &Long &20.96 &16.94 &19.66 &19.19 \\
Story Completion &Target &20.47 &18.52 &19.06 &19.35 \\
Story Completion & Lang  &19.71 &14.17 &20.24 &18.04 \\  
Plot Outline &None &10.93 &15.19 &7.298 &11.14 \\
Plot Outline &Long &17.78 &15.94 &17.64 &17.12 \\
Plot Outline &Target &15.53 &16.90 &16.81 &16.41 \\
 Plot Outline & Lang &18.10 &18.46 &18.60 &18.39 \\  
3Act Structure &None &22.70 &19.80 &22.36 &21.62 \\
3Act Structure&Long &22.09 &21.15 &23.16 &22.13 \\
3Act Structure & Target &20.99 &20.14 &24.63 &21.92 \\
 3Act Structure & Lang &{22.61} &{21.16} &{24.34} &{22.70} \\
\bottomrule
\end{tabular}
}
\caption{SentencePiece-ROUGE between generated and reference  story
  for PaLM with different plan variants and sory prefixes. None is a shorthand for the prefix
  ``Story:'', Long abbreviates prefix ``A native
  \textless{}tgt\_language\textgreater~speaker would write the story
  as:'', Target is the translation of the prefix ``Story:'' in the
  target language, and Lang refers to
  ``\textless{}tgt\_language\textgreater~story:''.}
\label{tab:spROUGE:full}
\end{table}

\begin{table*}[t]
\resizebox{\textwidth}{!}{%
\begin{tabular}{@{}ll@{}}
  \toprule
  \multicolumn{1}{c}{Entities}  & \multicolumn{1}{c}{Story Completion}   
\\ \midrule
\begin{tabular}[c]{@{}l@{}}Una volta, c'era \colorbox{lime}{una mucca, un cane e una capra} che andarono a visitare \\un altro villaggio. Decisero di prendere un autobus. Il cane era un po' \\preoccupato, ma la capra era molto felice e disse al cane: "Non preoccuparti!\\ La gente sarà gentile con noi." Il cane non era sicuro, ma la capra era determinata. \\Così tutti e tre i loro amici salirono sull'autobus. Quando \colorbox{lime}{il conducente} dell'autobus\\ li vide salire, cominciò a ridere. "Siete voi tre \colorbox{lime}{amici} che vi siete persi?" Disse \\ \colorbox{lime}{il conducente}. "Dove state andando?" \colorbox{lime}{La capra}, che era più intelligente \\dei \colorbox{lime}{suoi amici}, rispose: "Stiamo andando a visitare un altro villaggio." \colorbox{lime}{Il conducente}\\ dell'autobus si mise a ridere di nuovo. "Bene, amici! Non ho mai visto animali \\sull'autobus! Siete molto divertenti!" Disse.\colorbox{lime}{Il cane} non era sicuro. \colorbox{lime}{La capra} rispose:\\ "Abbiamo pagato abbastanza soldi per prendere l'autobus."\\ <...>\end{tabular} & \begin{tabular}[c]{@{}l@{}}  \colorbox{lime}{Capra, Cane e Mucca erano grandi amici. Un giorno, presero un taxi per} \\\colorbox{lime}{andare a fare un giro.} Capra sedeva sul sedile posteriore, mentre Cane e \\Mucca sedevano sul sedile anteriore. Quando il taxi passava attraverso \\ il villaggio, le persone dicevano: "Guardate! Guarda quel taxi! Che strano!"\\ \colorbox{lime}{Capra, Cane e Mucca} andarono a visitare un amico che viveva in una \\fattoria vicina. Quando il taxi arrivò alla fattoria, gli animali scesero \\e salutarono il loro amico. La fattoria era piena di erba fresca,\\ così Cane, Capra e Mucca decisero di fermarsi un po'.\\ Il taxi rimase fermo nella fattoria per sei ore.\\ Intanto, i tre amici giocarono e giocarono e giocarono.\\ <...>\end{tabular}\\ \midrule 
 \multicolumn{1}{c}{Plot Outline}  & \multicolumn{1}{c}{3Act Structure} \\ \midrule 
\begin{tabular}[c]{@{}l@{}}\colorbox{lime}{Una capra} voleva andare a
  visitare un suo amico. \\Aveva un problema. Non aveva idea di dove
  fosse il suo amico.\\ "Posso andare a visitare il mio amico?" Chiese
  alla \colorbox{lime}{mucca}. \\"No," rispose la mucca. "Ma tu sai
  dove si trova."\\ Disse la capra. "Sì," rispose la mucca. "Ma non ho
  tempo. Devo dormire." \\     "Posso andare a visitare il mio amico?"
  \\Chiese alla capra \colorbox{lime}{al cane}. "No," rispose il
  cane. "Ma tusai dove si trova." \\Disse la capra. "Sì," rispose il
  cane. "Ma non ho tempo. Devo giocare." \\ <...>\end{tabular}
  & \begin{tabular}[c]{@{}l@{}}\colorbox{lime}{Una volta, la mucca, la
        capra e il cane viaggiarono insieme sul
        taxi.}\\  <...>\\ Chiese la mucca. "Se non pagherai,\\ il
      conducente del taxi chiamarà la polizia e sarai arrestata.\\ Non
      puoi semplicemente scappare!" \colorbox{lime}{La capra non
        ascoltò e corse via. Il cane}\\ \colorbox{lime}{si precipitò verso il taxi per chiedere il resto all'autista.}\\ \colorbox{lime}{La mucca attraversò la strada con calma,}\\ \colorbox{lime}{perché aveva pagato il giusto prezzo.}\\ \colorbox{lime}{La capra uscì correndo dal taxi e scappò via.}\\ La capra non pagò il taxi. La capra fuggì.\\ \colorbox{lime}{La mucca attraversò la strada con calma.}\\ \colorbox{lime}{Il cane tornò al taxi.}\end{tabular} \\  
 \bottomrule
\end{tabular}%
}
\caption{Abridged examples of stories generated in Italian with different plans for story ``Goat, Dog and Cow'' (story ID: 4). 
Highlighted excerpts correspond to the events and/or characters
mentioned in the input plan.}
\label{tab: sample stories}
\end{table*}

\section{Qualitative Evaluation}
\label{sec:qual-eval}

Table~\ref{tab: sample stories} shows four abridged stories generated
in Italian based on different plans. We observe that all of them have
the same \emph{main characters} as prescribed in their respective
plans (i.e., the goat, the dog and the cow). A secondary character,
the taxi driver, is mentioned in stories based on entities and
three-acts but is absent when the model is given a plot outline or
asked to perform story completion. We hypothesise this is because the
taxi driver is not explicitly mentioned in the latter two plan
instantiations.  There are no new characters added to the story,
except in the case of plot outline where two novel characters are
present, namely a sheep and a chicken, who the main characters
address.

All stories convey the correct \emph{setup} (i.e., travelling between
places), However, the story based on three-acts conveys more detail,
while others make some mistakes, for instance, the characters use the
wrong means of transportation.
Most stories have a consistent \emph{flow} of events and final
resolution with the exception of the story based on the plot outline
which is repetitive (the main character repeatedly asks the same
questions). This indicates that coarse outlines might yield less
coherent stories, which are less likely to resolve into an ending. No
matter what plan is used, the stories are consistently
grammatical.  We further observed that the stories rarely use
pronouns, which are a diagnostic of locally coherent
discourse \cite{Grosz:ea:95}.


Our plans contain information which can be used in different parts of
the story. The highlights in Table~\ref{tab: sample stories} show
spans in the story which are directly copied from the plan. As can be
seen, entities are interspersed throughout (both as subjects and
objects) while story completion is most useful in guiding the
beginning of the story (the remainder is being generated
creatively). The plot outline is also less constraining leading to
more creative but less coherent stories. The PaLM follows the three-act
plan closely, drawing information  throughout.

\end{document}